\documentclass[10pt,twocolumn,letterpaper]{article}

\usepackage{cvpr}
\usepackage{times}
\usepackage{epsfig}
\usepackage{graphicx}
\usepackage{amsmath}
\usepackage{amssymb}
\usepackage{xcolor}
\usepackage{booktabs}
\usepackage{multirow}
\usepackage{gensymb}
\usepackage{xspace}
\usepackage{epsfig}
\usepackage{color}
\definecolor{ForestGreen}{rgb}{0.0, 0.5, 0.0}
\definecolor{DarkRed}{rgb}{0.7, 0.0, 0.0}
\definecolor{DarkMagenta}{rgb}{0.55, 0.0, 0.55}
\definecolor{DarkOrange}{rgb}{0.8, 0.25, 0.0}

\newcommand{\comment}[1]{}
\newcommand{\imgspace}[0]{\hspace{-0.3cm}}

\usepackage[pagebackref=true,breaklinks=true,letterpaper=true,colorlinks,bookmarks=false]{hyperref}

\cvprfinalcopy 

\def\cvprPaperID{5417} 

\ifcvprfinal\fi
\begin{document}

\title{An Annotation Saved is an Annotation Earned: \\ Using Fully Synthetic Training for Object Instance Detection}

\author{Stefan Hinterstoisser, Olivier Pauly\thanks{equal contribution}, Hauke Heibel \footnotemark[1], Martina Marek, Martin Bokeloh \footnotemark[1] \\
\footnotetext{equal contribution}
Google Cloud AI\\
Erika-Mann-Strasse 33, 80636 Munich, Germany\\
{\tt\small \{hinterst,olivierpauly,haukeheibel,mmmarek,mbokeloh\}@google.com}
}

\maketitle

\begin{abstract}
Deep learning methods typically require vast amounts of training data to reach their full potential. While some publicly available datasets exists, domain specific data always needs to be collected and manually labeled, an expensive, time consuming and error prone process. Training with synthetic data is therefore very lucrative, as dataset creation and labeling comes for free. We propose a novel method for creating purely synthetic training data for object detection. We leverage a large dataset of 3D background models and densely render them using full domain randomization. This yields background images with realistic shapes and texture on top of which we render the objects of interest. During training, the data generation process follows a curriculum strategy guaranteeing that all foreground models are presented to the network equally under all possible poses and conditions with increasing complexity. As a result, we entirely control the underlying statistics and we create optimal training samples at every stage of training. Using a set of 64 retail objects, we demonstrate that our simple approach enables the training of detectors that outperform models trained with real data on a challenging evaluation dataset.

\end{abstract}

\section{Introduction}
\begin{figure}[ht]
\begin{center}
\begin{tabular}{cc}
\imgspace
\includegraphics[width=0.5\linewidth]{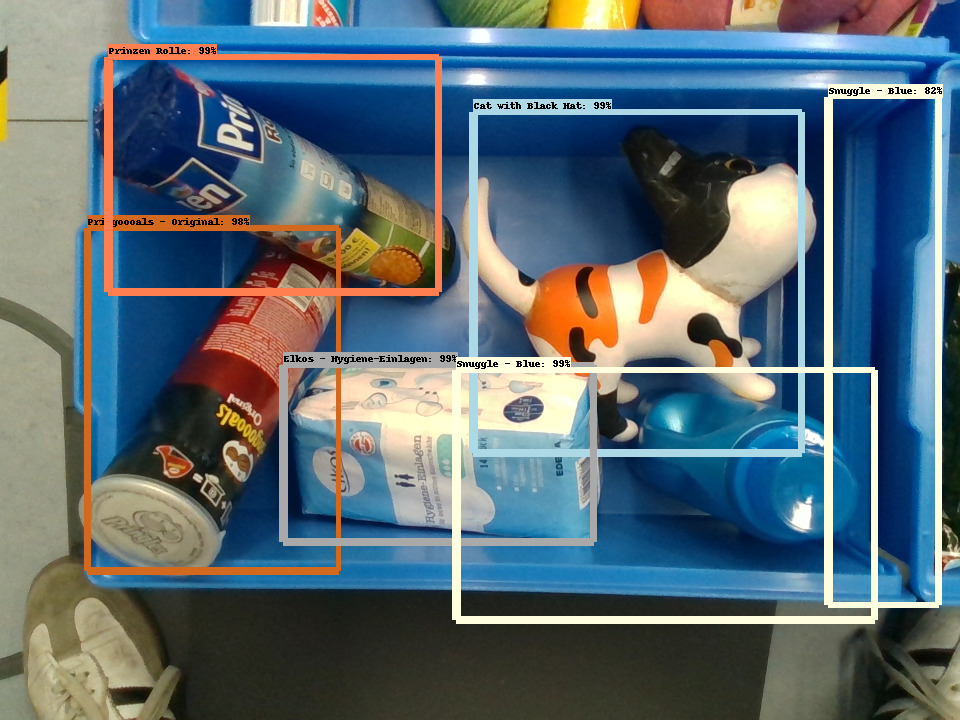} & \imgspace
\includegraphics[width=0.5\linewidth]{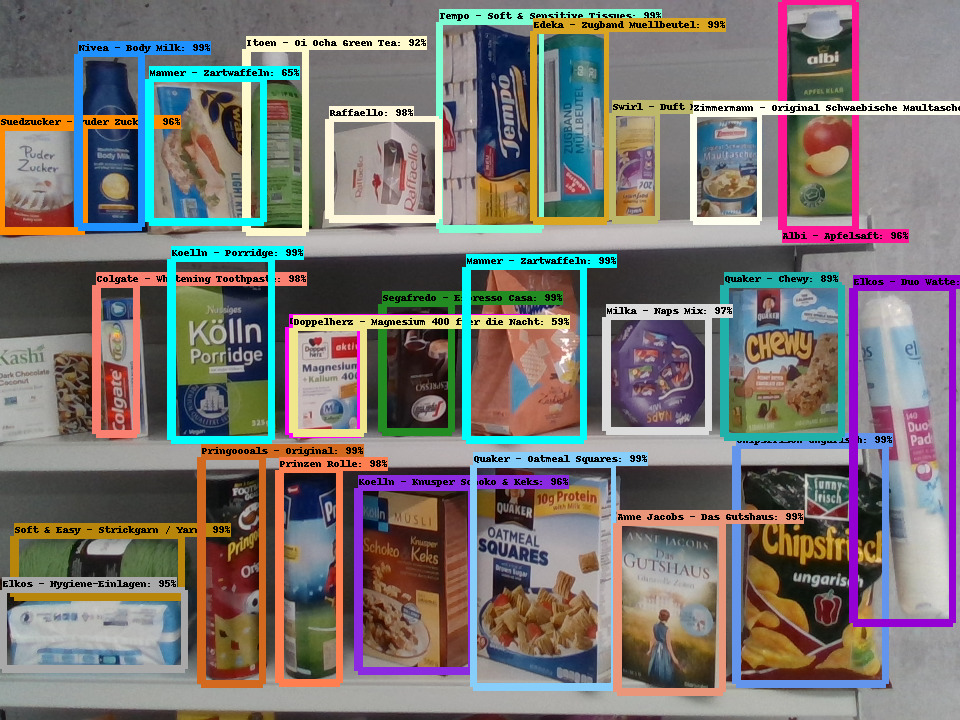} \\
\imgspace
\includegraphics[width=0.5\linewidth]{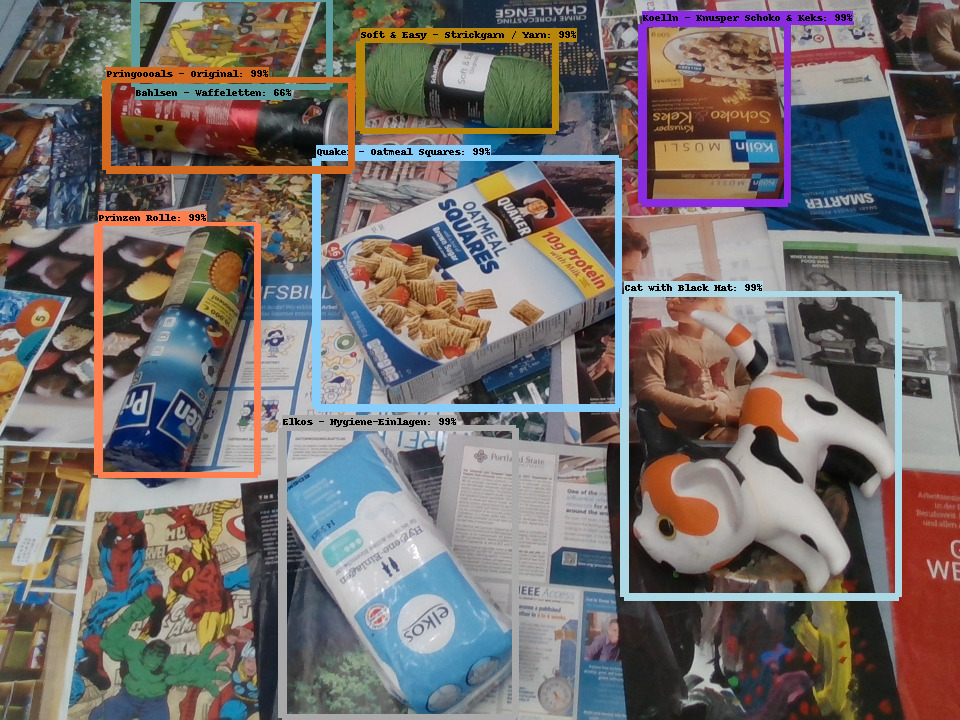} & \imgspace
\includegraphics[width=0.5\linewidth]{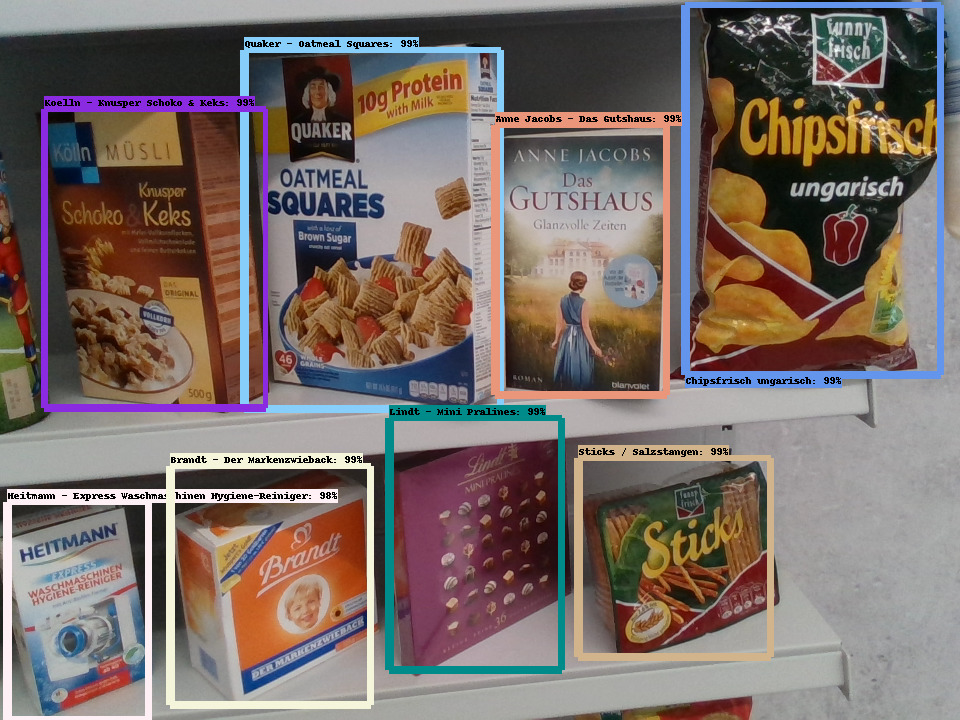} \\
\end{tabular}
\end{center}
\caption{\label{fig:teaser} Example results of Faster R-CNN \cite{faster_rcnn} trained on \textit{purely synthetic data} from 3D models. 
In this paper we introduce a novel approach for creating synthetic training data for object detection that generalizes well to real data. Our trained model is able to robustly detect objects under various poses, heavy background clutter, partial occlusion and illumination changes.}
\end{figure}

The capability of detecting objects in challenging environments is fundamental for many machine vision and robotics tasks. Recently, proposed modern deep convolutional architecture such as Faster R-CNNs~\cite{faster_rcnn},  SSD~\cite{ssd},  R-FCN~\cite{rfcn}, Yolo9000~\cite{redmon16} and RetinaNet~\cite{Lin2017b} have achieved very impressive results. However, the training of such models with millions of parameters requires a massive amount of labeled training data to achieve state-of-the-art results. Clearly, the creation of such massive datasets has become one of the main limitations of these approaches: they require human input, are very costly, time consuming and error prone.

\comment{
The capability of detecting objects in challenging environments is fundamental for many machine vision and robotics tasks. With the introduction of modern deep convolutional architectures such as Faster R-CNNs~\cite{faster_rcnn},  SSD~\cite{ssd},  R-FCN~\cite{rfcn}, Yolo9000~\cite{redmon16} and RetinaNet~\cite{Lin2017b} very impressive results have been achieved. However, the training of such models with millions of parameters requires a massive amount of labeled training data to achieve state-of-the-art results. Clearly, creating such massive datasets becomes one of the main bottlenecks for those approaches: it requires human input, and it is very costly, time consuming and error prone.
}

Training with synthetic data is very attractive because it decreases the burden of data collection and annotation. Theoretically, this enables generating an infinite amount of training images with large variations, where labels come at no cost. In addition, training with synthetic samples allow to precisely control the rendering process of the images and thereby the various properties of the dataset. However, the main challenge for successfully applying such approaches in practice still remains, i.e. how to bridge the so-called ``domain gap" between synthesized and real images. As observed in \cite{Tobin17}, methods trained on synthetic data and evaluated on real data usually result in deteriorated performance.

\comment{
To decrease the burden of data collection and annotation, training with synthetic data is very attractive. Theoretically, this permits to generate an infinite amount of training images with large variations, where labels come at no cost. In addition, it allows to precisely control the rendering process of the images and thus the various properties of the dataset. However, the main challenge for successfully applying such approaches in practice still remains, i.e. how to bridge the so-called ``domain gap" between synthesized and real images. As observed in \cite{Tobin17}, the transfer from synthetic to real domain usually results in deteriorated performance.
}

To address this challenge, several approaches have focused on improving the realism of training data~\cite{Gupta16,Alhaija17,Georgakis17,Varol17}, mixing synthetic and real data~\cite{Dwibedi17,Georgakis17,Rad17c}, leveraging architectures with frozen pre-trained feature extractors \cite{Hinterstoisser18, kehl2017iccv, Rajpura17}, or using domain adaptation or transfer learning as in \cite{Rozantsev17,bousmalis2016domain,ganin2016domain}. 

``Domain Randomization" as introduced in \cite{Tobin17} is another strategy to narrow the gap between real and synthetic data. The authors hypothesized that high randomization of the synthesis process yields better generalization as reality is seen by the trained models as a mere instance of the larger domain space it was trained on. They showed promising first results with a few objects in simple scenarios. More recently, this idea was extended with the addition of real background images mixed with partial domain randomized scenes~\cite{Tremblay18,Prakash18}, and further improved through photo-realistic rendering~\cite{tremblay2018corl}. While those approaches provided impressive results, the main drawback still remains i.e. their dependence on real data.

\comment{
``Domain Randomization" as introduced in \cite{Tobin17} is another strategy to narrow the gap between real and synthetic data. In this work, authors hypothesized that high randomization of the synthesis process permits to achieve better generalization as reality is seen by the trained models as a mere instance of the larger domain space it was trained on. \cite{Tobin17} showed promising first results with a few objects in simple scenarios. More recently, this idea was extended with the addition of real background images mixed with partial domain randomized scenes~\cite{Tremblay18,Prakash18}, and further improved through photo-realistic rendering~\cite{tremblay2018corl}. While those approaches provided impressive results, the main drawback still remains i.e. their dependence on real data.
}

In this paper, we introduce a novel way to create purely synthetic training data for object detection. We leverage a large dataset of 3D background models which we densely render in a fully domain randomized fashion to create our background images. Thus, we are able to generate locally realistic background clutter which makes our trained models robust to environmental changes. On top of these background images, we render our 3D objects of interest. During training, the data generation process follows a curriculum strategy which ensures that all foreground models are presented to the network equally under all possible poses with increasing complexity. Finally, we add randomized illumination, blur and noise.

\comment{
In this paper, we introduce a novel way to create purely synthetic training data for object detection. We leverage a large dataset of 3D background models which we densely render in a fully domain randomized fashion to create our background images. Thus, we are able to generate locally realistic background shapes via the shapes of the 3D models and to control the background clutter which makes our trained models robust to environmental changes. On top of these background images, we render our 3D objects of interest. During training, the data generation process follows a curriculum strategy that ensures that all foreground models are presented to the network equally under all possible poses with increasing complexity. Finally, we add randomized light, blur and noise.
}

Our approach doesn't require complex scene compositions as in~\cite{tremblay2018corl,Gupta16,Alhaija17,Georgakis17,Varol17}, difficult photo-realistic image generation as in~\cite{tremblay2018corl,Gupta16,Alhaija17} or real background images to provide the necessary background clutter~\cite{Hinterstoisser18,kehl2017iccv,Rajpura17,Tremblay18,Prakash18,tremblay2018corl}, and scales very well to a large number of objects and general detection capabilities. 

To the best of our knowledge we are the first to present such a purely synthetic method for generating training data for object instance detection that outperforms models trained on real data. Furthermore, we demonstrate experimentally the benefits of curriculum strategy versus random pose generation. We also show that generated images should ideally be composed of synthetic content only and that the whole background image should be filled with background clutter. Finally, we perform thorough ablation experiments to highlight the contributions of the different components of our pipeline.

In the remainder of the paper we first discuss related work, describe our pipeline for generating synthetic images, demonstrate the usefulness of fully synthetic data, and detail our experiments and conclusions.

\section{Related Work}
A common approach to improve detection performance is to extend a real training dataset by adding synthetic data. 
For instance, \cite{Su15, Dwibedi17,Georgakis17} train a single network on such a mixed dataset. 
While these methods demonstrate a significant improvement over using real data only, they still require at minimum real domain-specific background images as in \cite{Su15}. 

\cite{Dwibedi17,Georgakis17} follow an image composition approach to create synthetic images by combining cut out objects from different images. These approaches have the benefit of using data from the same domain, as the cut out objects are copies of real images, and as such, they closely match the characteristics of the real world. The main limitation of these approaches is that they require performing the cumbersome process of capturing images of the objects from all possible viewpoints and mask them. In particular, these methods can't produce images from different views or different lighting conditions once the object training set is fixed. This is a clear limitation.

Other lines of work utilize photo-realistic rendering and realistic scene compositions to overcome the domain gap by synthesizing images that match the real world as close as possible \cite{Gupta16,JohnsonRoberson16,Richter_2016_ECCV, Mitash17, Alhaija17, Georgakis17,Varol17, MovshovitzAttias16}. 
While these methods have shown promising results they face many hard challenges. First, producing photo-realistic training images requires sophisticated rendering pipelines and considerable CPU/GPU resources. 
Second, realistic scene composition is a hard problem on its own usually done by hand. 
Third, modern rendering engines used for creating synthetic scenes heavily take advantage of the human perception system to fool the human eye. However, these tricks do not necessarily work on neural networks and thus require more effort to bridge the domain gap.

Following their success for image generation, Generative Adversarial Networks~(GANs) have been used in~\cite{Shrivastava16,Bousmalis17} to further bridge the domain gap. However, such approaches bring substantial additional complexity as they are difficult to design and train. To the best of our knowledge they have not been applied to detection tasks yet.


Another line of work utilizes domain adaptation or transfer learning \cite{Rozantsev17,bousmalis2016domain,ganin2016domain,Inoue18} to bridge the domain gap between the synthetic and real domain. This can be achieved by coupling two predictors, one for each domain, or by combining the data from two domains. Domain adaptation and transfer learning have applications far beyond the transfer from synthetic to real data. Still, they require a significant amount of real data.

Our method falls into the category of domain randomization \cite{Tobin17, Tremblay18, tremblay2018corl, Prakash18, Borrego18}. The basic idea is to alter the simulated data with non-realistic changes so that reality seems to be just a variation. \cite{Tobin17} introduced the concept of domain randomization to overcome the domain gap. They use non-realistic textures for rendering synthetic scenes to train an object detector which generalizes to the real world. 
In another line of work, \cite{tremblay2018corl} combines domain randomization and photo-realistc rendering. They generate two types of data: First, synthetic images with random distractors and variations that appear unnatural with real photographs as background as introduced in \cite{Tremblay18}, and second, photo-realistic renderings of randomly generated scenes using a physics engine to ensure physical plausibility. The combination of these two types of data yields great improvement over only one source of data and allows the network to generalize to unseen environments. 
\cite{Prakash18} uses structured domain randomization, which allows the network to take context into account. In the context of structured environments such as street scenes, this yields state-of-the-art results, but is not applicable to scenarios like picking an item out of a box where there are no clear spatial relationships between the location of the different objects.
\section{Method} \label{sec:method}
\begin{figure*}[ht]
\begin{center}
\includegraphics[trim={0.0cm 9.7cm 0cm 0cm},clip=true,width=1\linewidth]{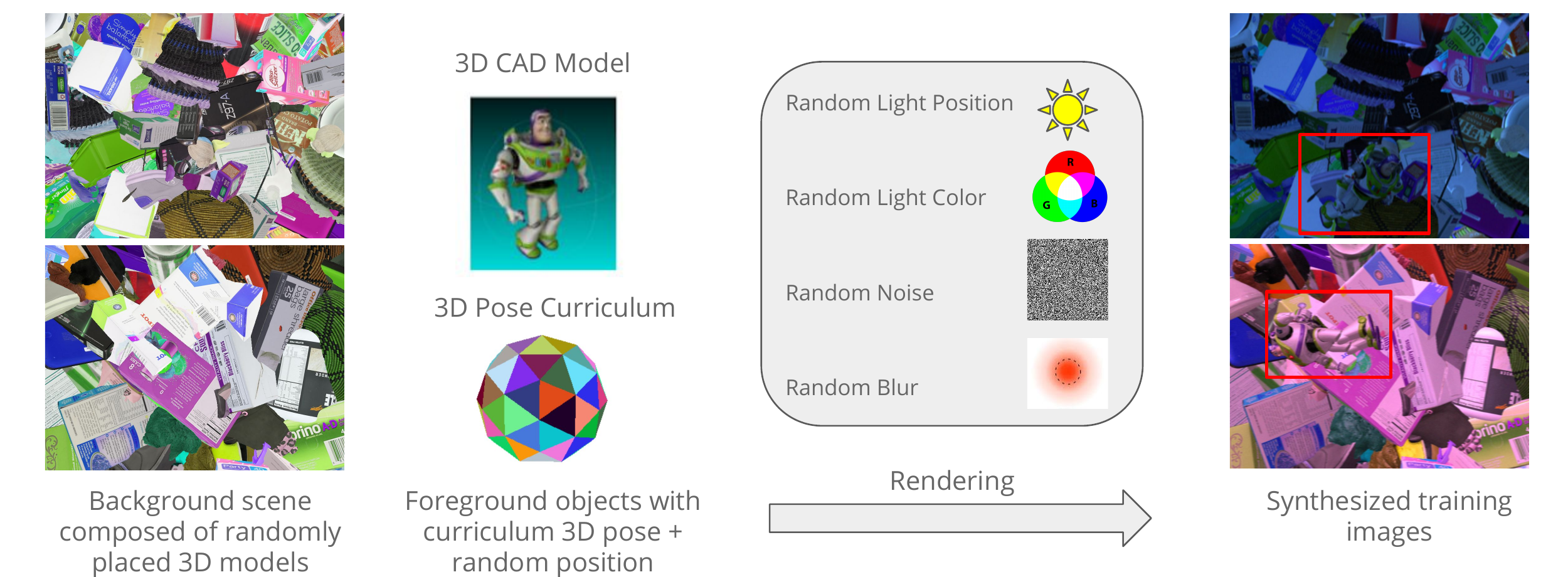} 
\end{center}
\caption{\label{fig:pipeline} Our synthetic data  generation pipeline.  
For each training image we generate a background scene by randomly placing 3D models from a background object database until each pixel in the resulting image would be covered (see Section \ref{sec_bg_layer}).
Then, we add one or many foreground objects to the scene; each object is randomly positioned in the image but follows a deterministic schedule for rotation and scale (see curriculum strategy in Section \ref{sec_fg_layer}).
Finally, we render the scene using simple Phong illumination~\cite{Phong75} with a randomly placed light source with a random light color, followed by adding random noise to the image and random blur.
We also compute a tightly fitting bounding box using the object's 3D model and the corresponding pose.}
\end{figure*}

In this section, we present our pipeline for generating synthetic training data as shown in Fig.~\ref{fig:pipeline}. As opposed to previous methods~\cite{Dwibedi17,Georgakis17,Rad17c}, we do not try to diminish the domain gap by mixing synthetic and real images but create purely synthesized training samples. Each training sample is generated by blending three image layers - a purely synthetic background layer, a foreground object layer built following a curriculum strategy and finally a last layer containing occluders. 

Since we are dealing with object instance detection and are interested in rendering our objects geometrically correct, we make use of the internal camera parameters, i.e. focal lenth and principal point. To gain additional robustness, we allow for slight random variations of these parameters during training. 

In the remainder of this section, we will describe in detail how we create each of these layers and the underlying principles which guided the design of the rendering pipeline.

\subsection{Background Layer Generation} \label{sec_bg_layer}
The background generation method is designed following three principles: maximize background clutter, minimize the risk of showing a network the same background image twice, and create background images with structures being similar in scale to the objects in the foreground layer. Our experiments indicate that these principles help to create training data which allows networks to learn the geometric and visual appearance of objects while minimizing the chances of learning to distinguish synthetic foreground objects from background objects simply from different properties like e.g. different object sizes or noise distributions. 

The background layer is generated from a dataset of 15k textured 3D models, which is disjoint from the foreground object dataset. 
All 3D background models are initially de-meaned and scaled such that they fit into a unit sphere.

The background layer is created by successively selecting regions in the background where no other object has been rendered, and rendering a random background object onto this region. 
Each background object is rendered with a random pose and the process is repeated until the whole background is covered with synthetic background objects.

Key to the background generation is the size of the projected background objects, which is determined with respect to the size of the foreground object as detailed in~\ref{sec_fg_layer}.
Therefore, we generate a randomized isotropic scaling $\mathbf S$ which we apply to our unified 3D models before rendering them. 
We use the scaling to create objects such that the size of their projections to the image plane corresponds to the size of the average foreground object. 
More specifically, we compute a scale range $\mathcal{S} = \left[s_{min}, s_{max}\right]$ which represents the scales which can be applied to objects such that they appear within $[0.9,1.5]$ of the size corresponding to the average foreground object size. 
For each background image, we then create a random sub-set $\mathcal{S}_{bg} \subset \mathcal{S}$ to ensure that we do not only create background images with objects being uniformly distributed across all sizes, but also ones with primarily large or small objects. The isotropic scaling value $s_{bg}$ is now drawn randomly from $\mathcal{S}_{bg}$ such that background object sizes in the image are uniformly distributed.

For each background scene, we additionally convert each object's texture into HSV space, randomly change the hue value and convert it back to RGB to diversify backgrounds and to make sure that background colors are well distributed.

\subsection{Curriculum Foreground Layer Generation} \label{sec_fg_layer}

\begin{figure}[ht]
\begin{center}
\includegraphics[trim={0.3cm 1.5cm 6.0cm 0.4cm},clip=true,width=0.90\linewidth]{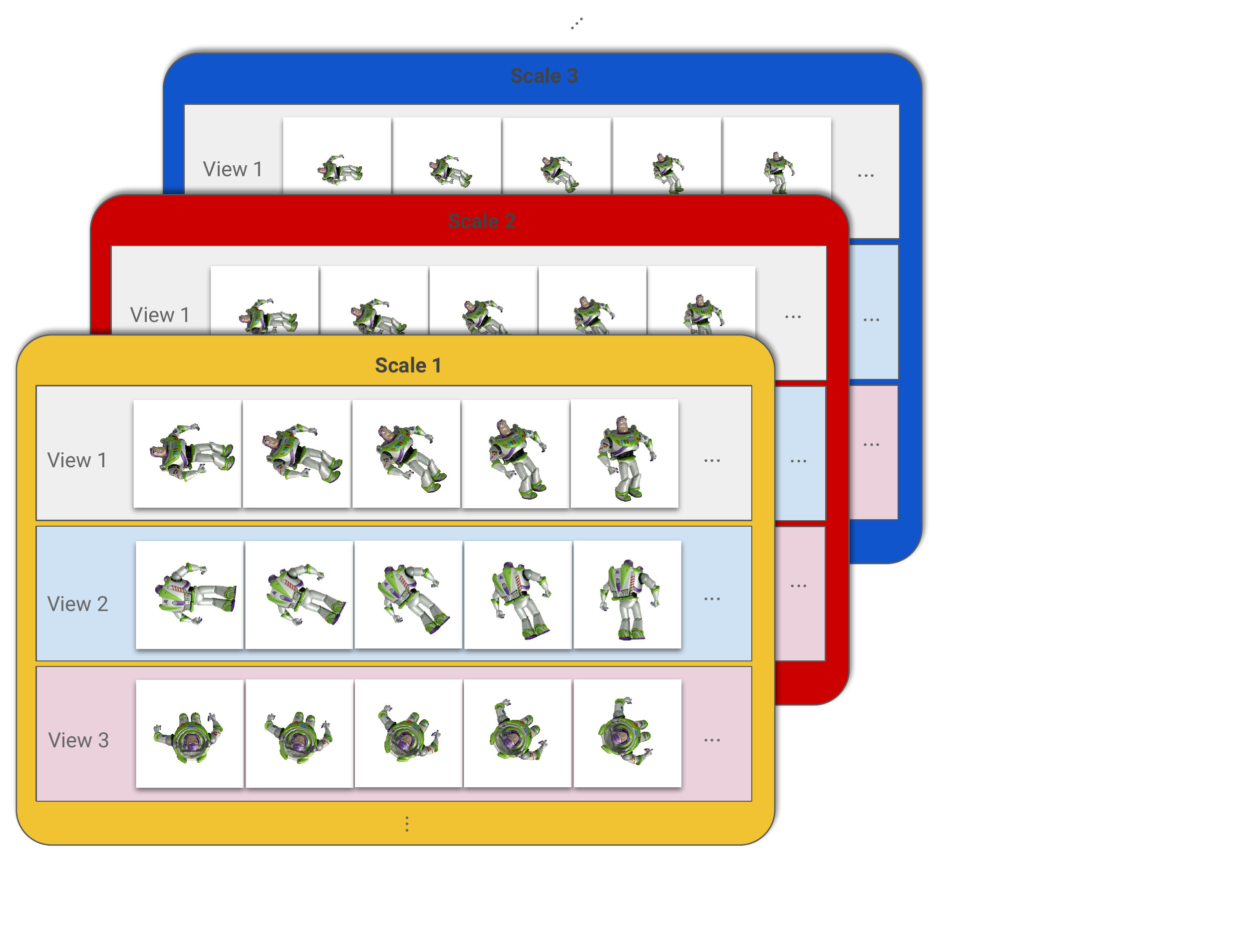} 
\caption{\label{fig:curriculum_example} Example curriculum for a single object. We show the object in the following order to the network: we start with the first scale and view and iterate through all in-plane rotations, followed by different out-of-plane rotations at the same scale. Once we have iterated through all in- and out-of-plane rotations, we proceed to the next scale in the same fashion.} 
\end{center}
\end{figure}

For each foreground object, we start by generating a large set of poses uniformly covering the pose space in which we want to be able to detect the corresponding object. To do so, we use the approach described in ~\cite{Hinterstoisser18} and generate rotations by recursively dividing an icosahedron, the largest convex regular polyhedron. This approach yields uniformly distributed vertices on a sphere and each vertex represents a distinct view of an object defined by two out-of-plane rotations. In addition to these two out-of-plane rotations, we also use equally sampled in-plane rotations. Furthermore, we sample the distance at which we render a foreground object inversely proportional to its projected size to guarantee an approximate linear change in pixel coverage of the projected object between consecutive scale levels.

Opposite to the background generation, we render the foreground objects based on a curriculum strategy (see Fig.~\ref{fig:curriculum_example}). This means that there is a deterministic schedule at which step each object and pose should be rendered:
\begin{enumerate}
\item We start with the scale that is closest to the camera and gradually move to the one that is farthest away. As a result, each object initially appears largest in the image, being therefore easier to learn for the network. As learning proceeds, the objects become smaller and more difficult for the network to learn.
\item For each scale, we iterate through all possible out-of-plane rotations, and for each out-of-plane rotation, we iterate through all in-plane rotations.
\item Once we have a scale, an out-of- and an in-plane rotation, we iterate through all objects, and render each of them with the given pose at a random location using a uniform distribution.
\item After having processed all objects, at all in- and out-of plane rotations, we move to the next scale level.
\end{enumerate}

For rendering, we allow cropping of foreground objects at the image boundaries up to $50\%$. In addition, we allow for overlap between each pair of foreground objects up to $30\%$.  For each object, we randomly try to place it $n=100$ times in a foreground scene. If it can't be placed within the scene due to violations of the cropping or overlap constraints we stop processing the current foreground scene and start with the next one. For the subsequent foreground scene, we start where we have left off the last scene.

\subsection{Occlusion Layer Generation}
We also generate an occlusion layer where we allow random objects from the background dataset to partially occlude the foreground objects. 
This is done by determining the bounding box of each rendered foreground object and by rendering a randomly selected occluding object at a uniform random location within this bounding box. 
The occluding object is randomly scaled such that its projection covers a certain percentage of the corresponding foreground object (in a range of $10\%$ to $30\%$ of the foreground object). 
The pose and color of the occluding object is randomized in the same way it is done for background objects. 

\subsection{Postprocessing and Layer Fusion}
Having the background, foreground and occlusion layer, we fuse all three layers to one combined image: 
the occlusion layer is rendered on top of the foreground layer and the result is rendered on top of the background layer. 
Furthermore, we add random light sources with random perturbations in the light color. Finally, we add white noise and blur the image with a Gaussian kernel where both, the kernel size and the standard deviation, are randomly selected. Thus, background, foreground and the occluding parts share the same image properties which is contrary to other approaches \cite{Hinterstoisser18, kehl2017iccv, Rajpura17,Tremblay18,Prakash18,tremblay2018corl} where real images and synthetic renderings are mixed.
This makes it impossible for the network to differentiate foreground vs. background merely on attributes specific to their domain.
In Fig.~\ref{fig:pipeline} we show some images generated with our method.

\section{Experiments}
\begin{figure*}[ht]
\begin{center}
\includegraphics[width=1\linewidth]{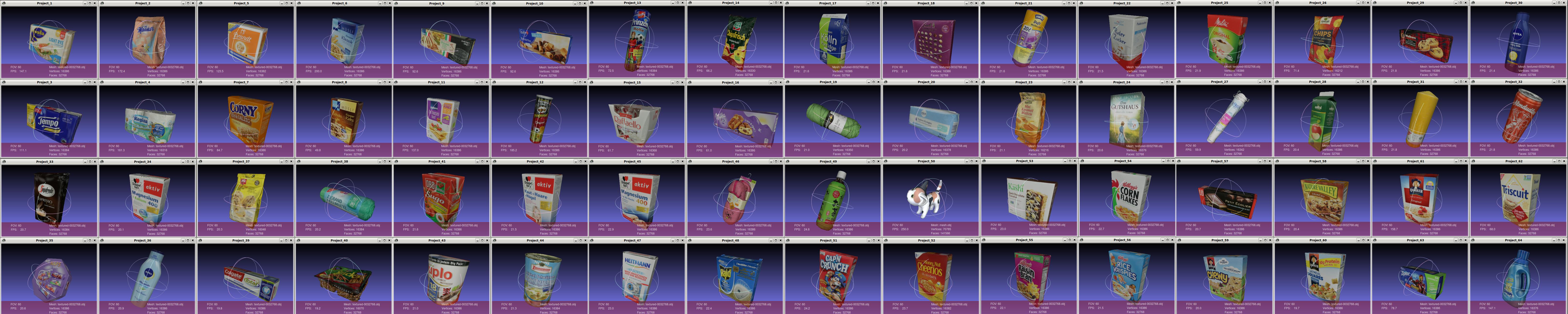} 
\end{center}
\caption{\label{fig:objects} The 64 objects of our training and evaluation dataset.}
\end{figure*}

\begin{figure*}
\begin{center}
\begin{tabular}{cccc}
\imgspace
\includegraphics[width=0.245\linewidth]{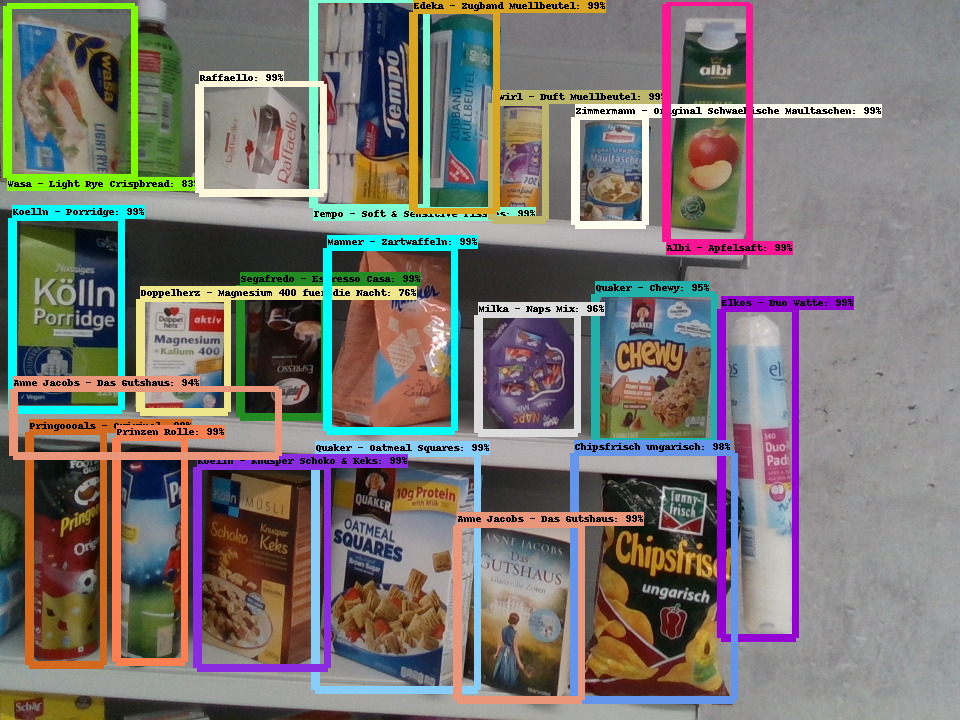} & \imgspace
\includegraphics[width=0.245\linewidth]{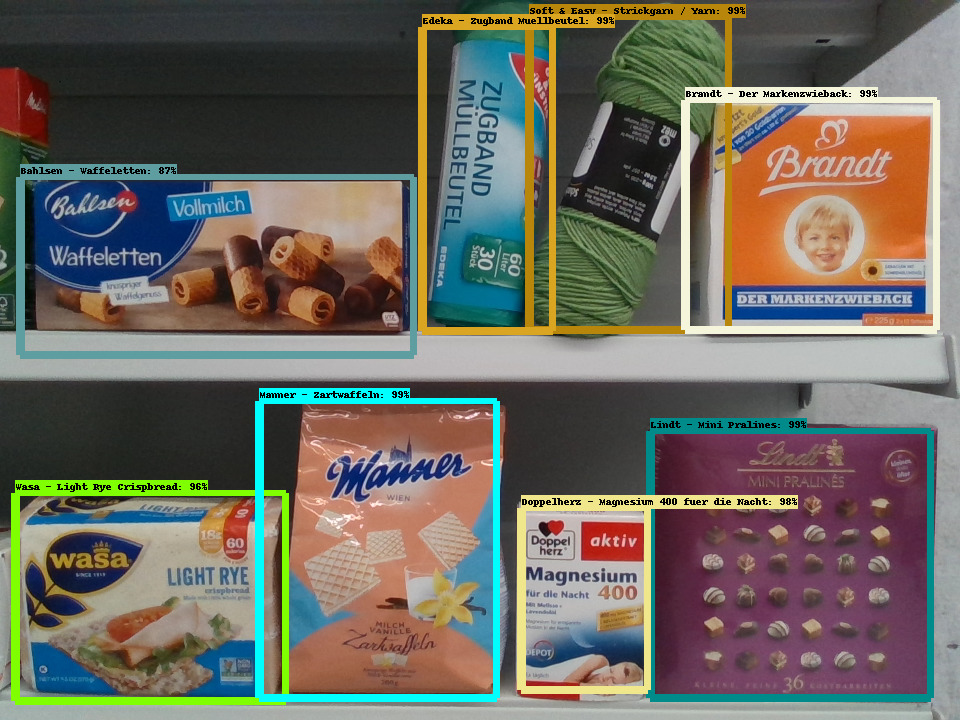} & \imgspace
\includegraphics[width=0.245\linewidth]{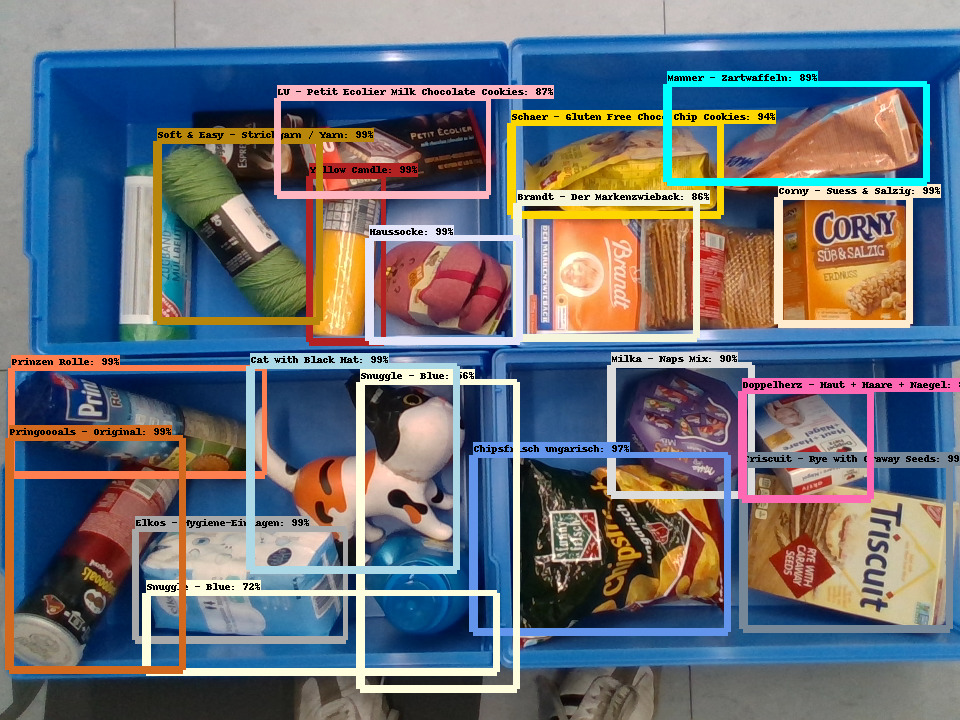} & \imgspace
\includegraphics[width=0.245\linewidth]{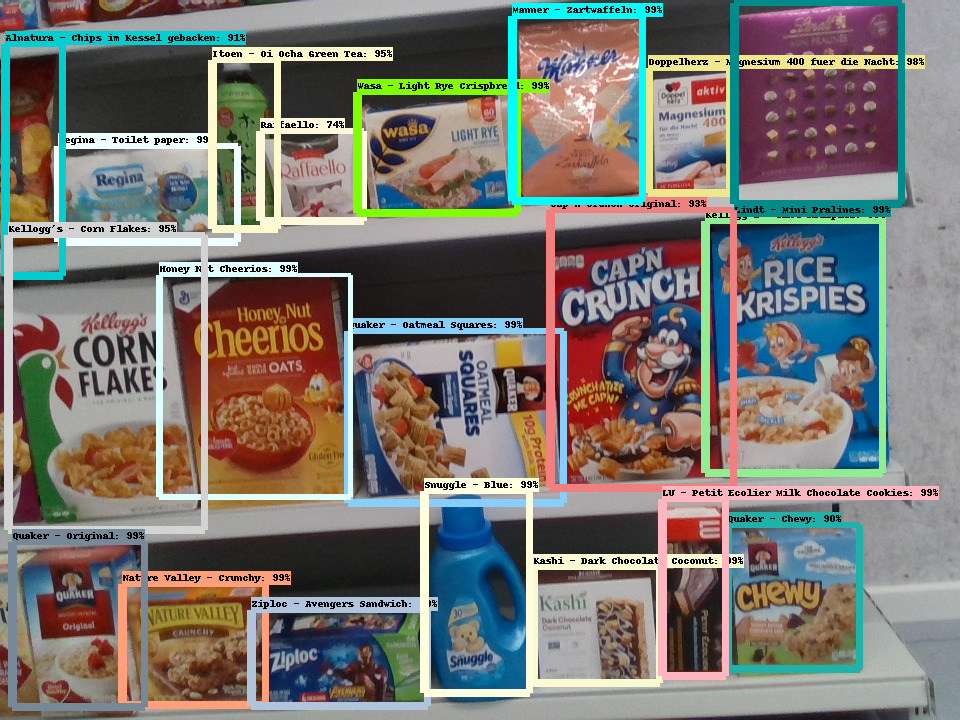} \\
\imgspace
\includegraphics[width=0.245\linewidth]{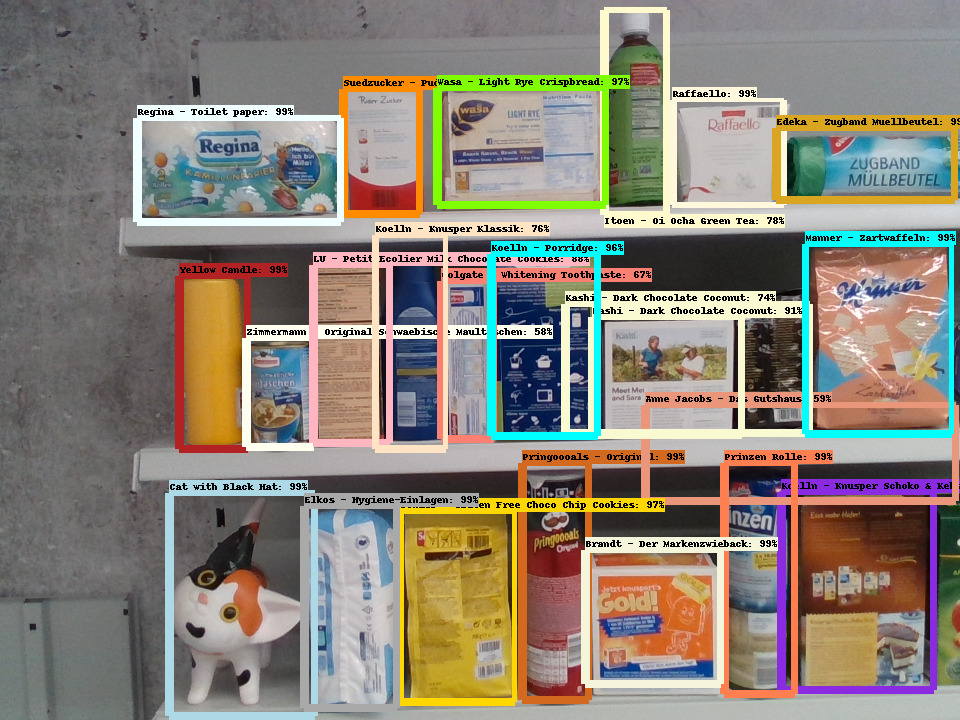} & \imgspace
\includegraphics[width=0.245\linewidth]{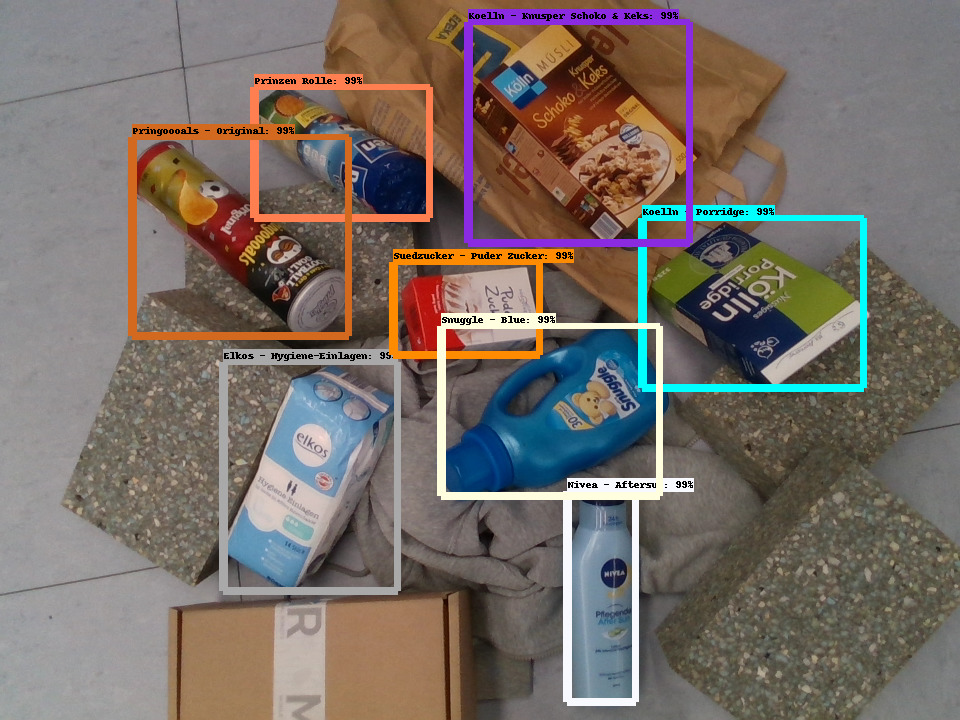} & \imgspace
\includegraphics[width=0.245\linewidth]{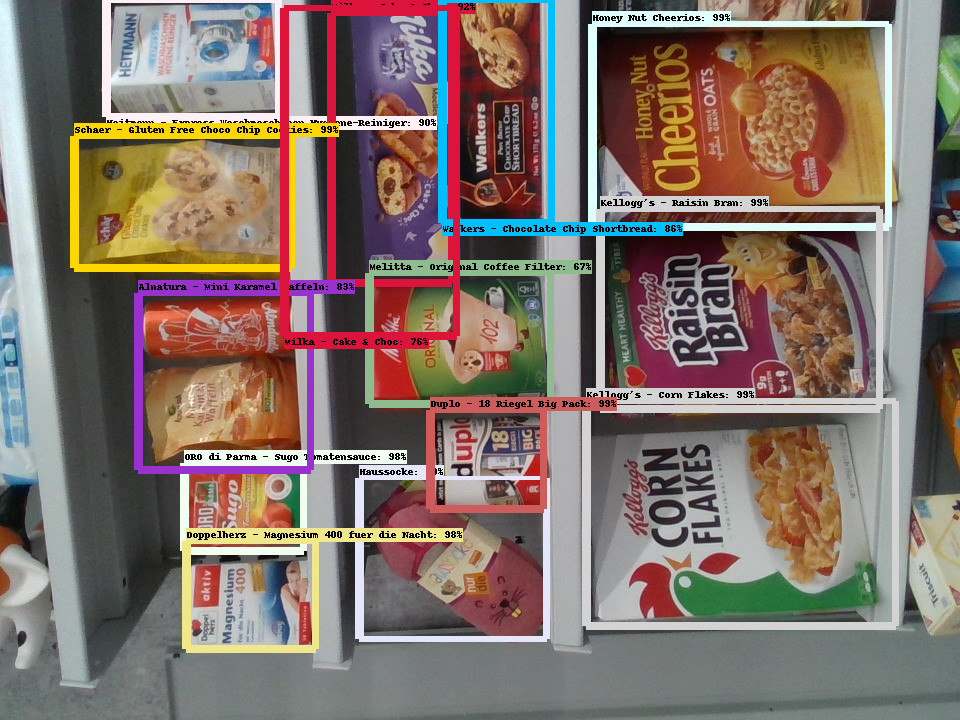} & \imgspace
\includegraphics[width=0.245\linewidth]{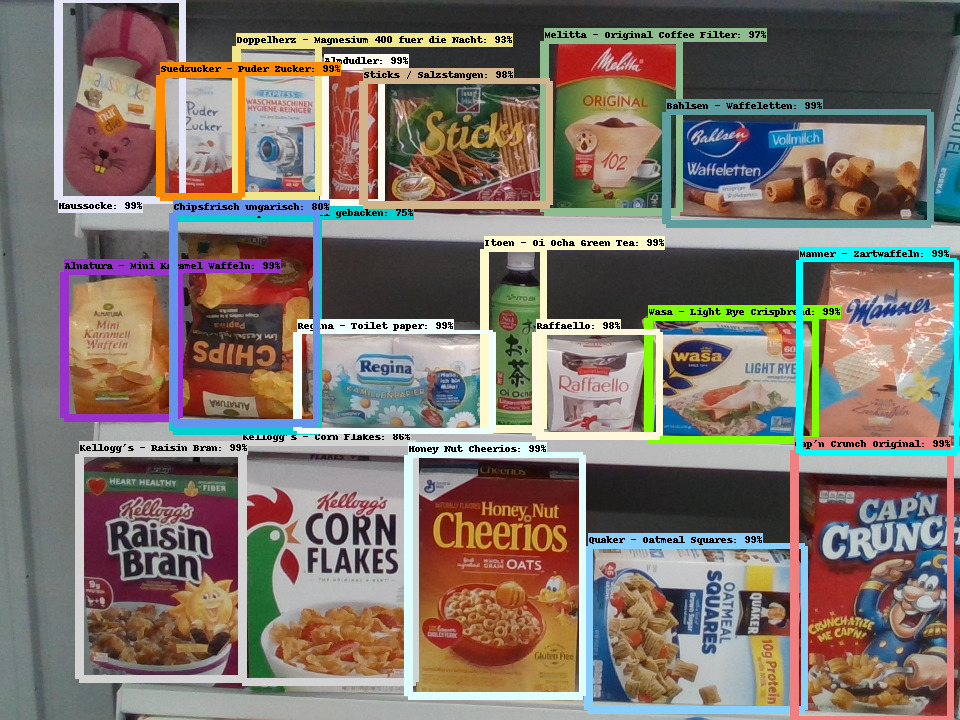} \\
\imgspace
\includegraphics[width=0.245\linewidth]{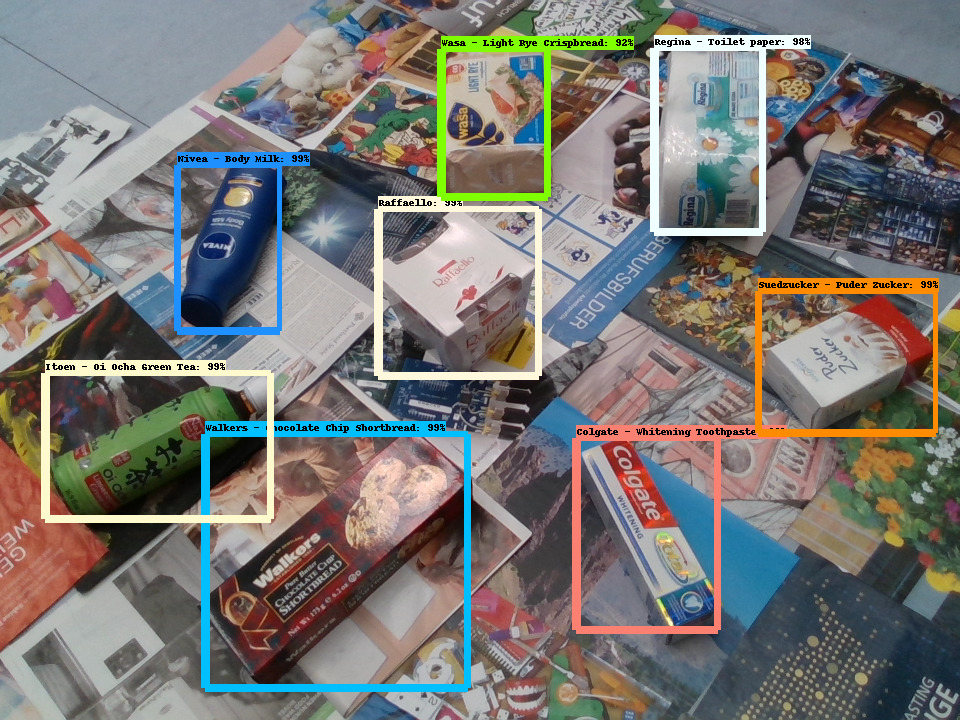} & \imgspace
\includegraphics[width=0.245\linewidth]{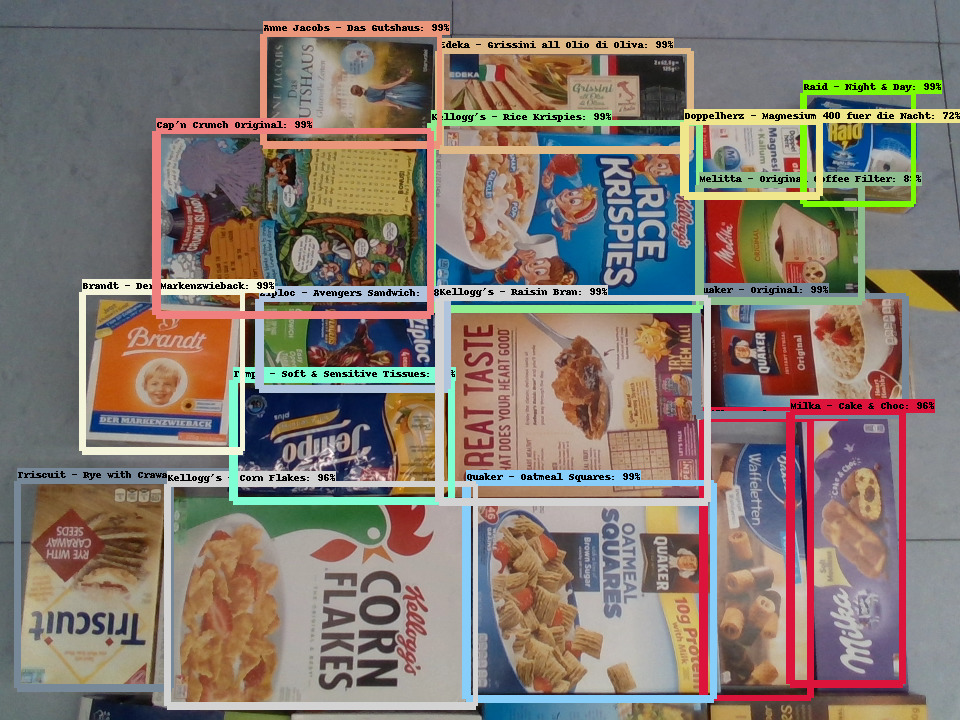} & \imgspace
\includegraphics[width=0.245\linewidth]{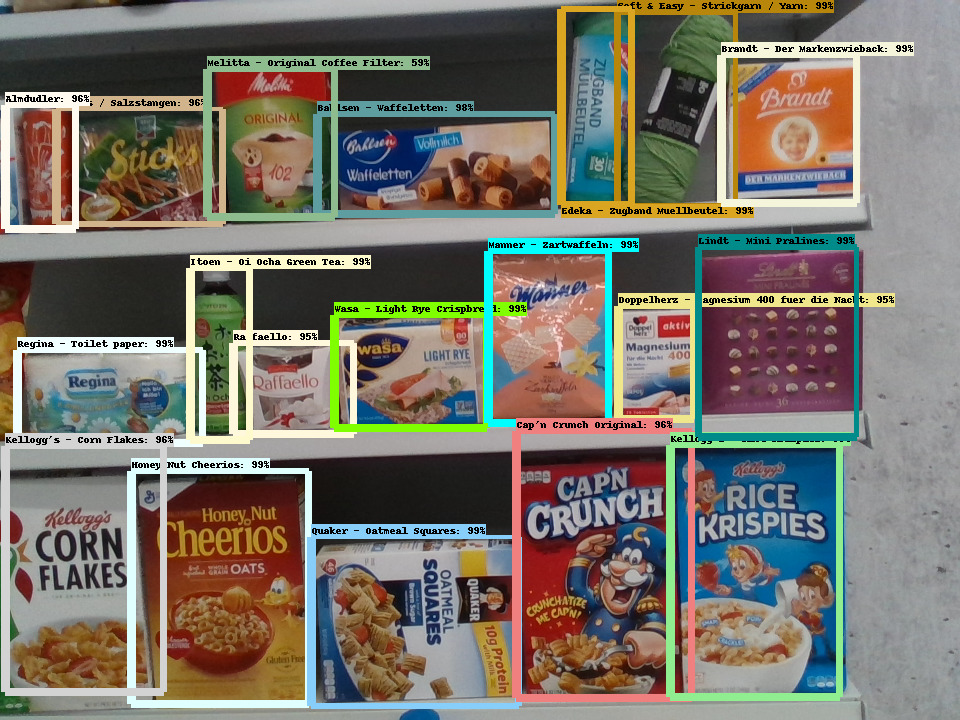} & \imgspace
\includegraphics[width=0.245\linewidth]{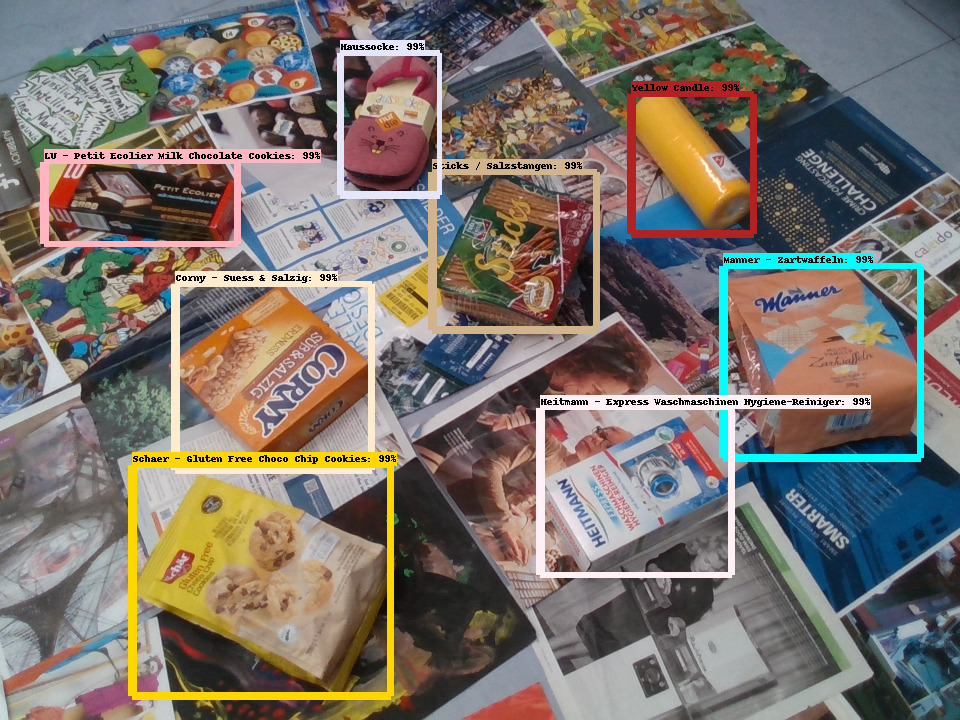} \\
\imgspace
\includegraphics[width=0.245\linewidth]{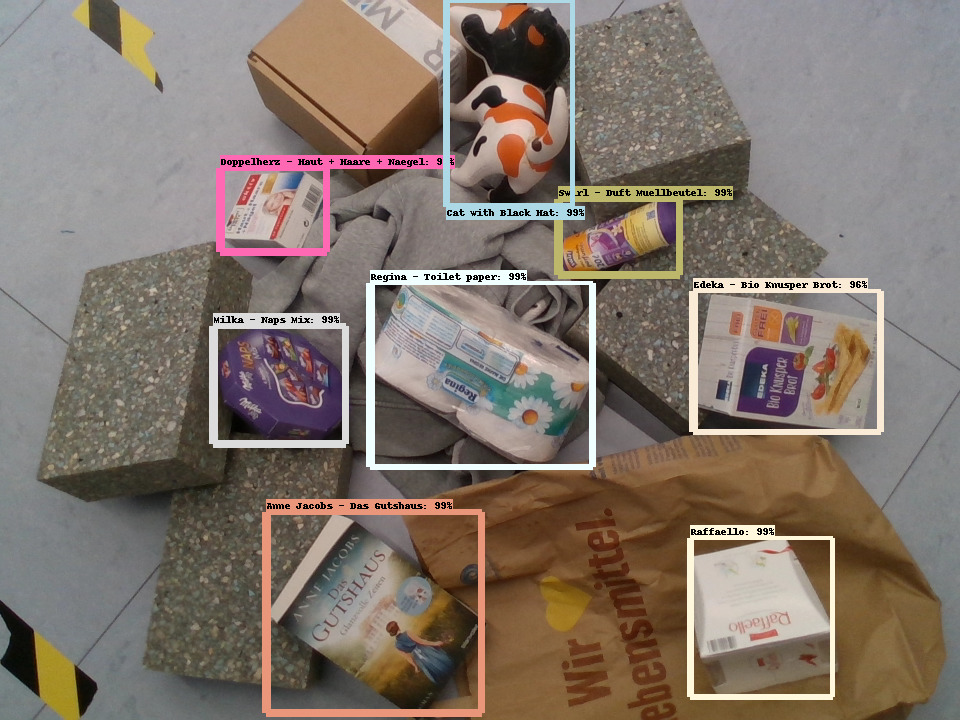} & \imgspace
\includegraphics[width=0.245\linewidth]{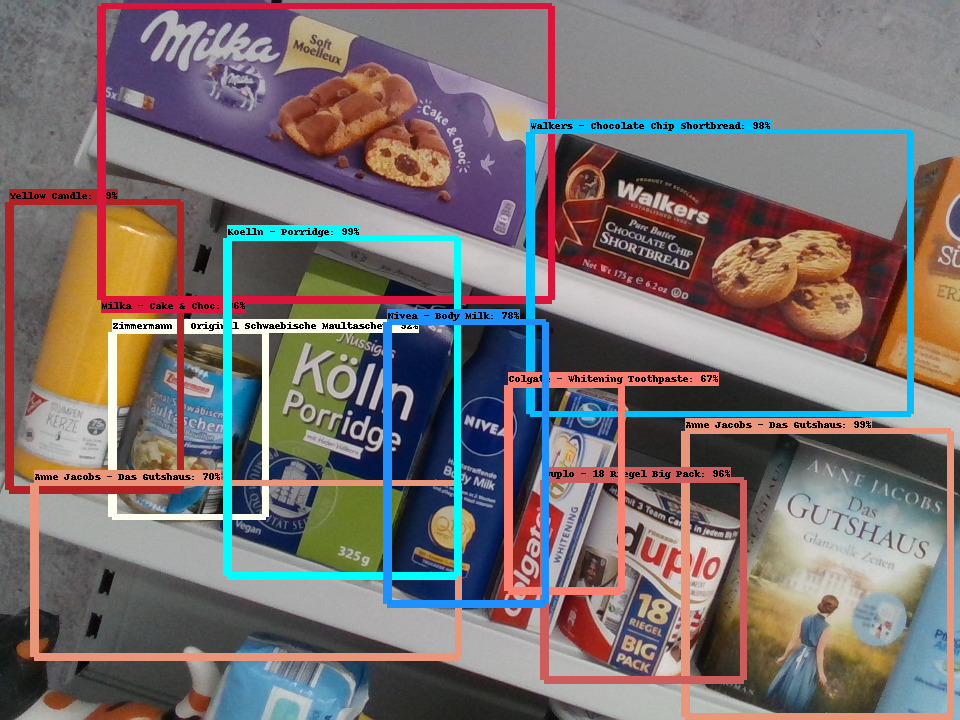} & \imgspace
\includegraphics[width=0.245\linewidth]{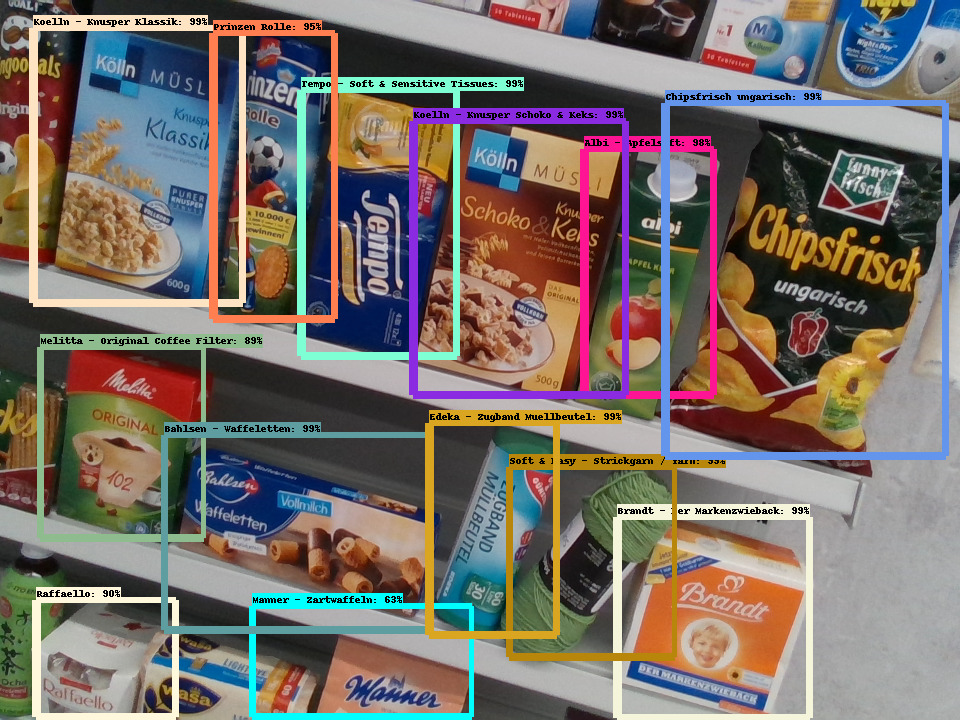} & \imgspace
\includegraphics[width=0.245\linewidth]{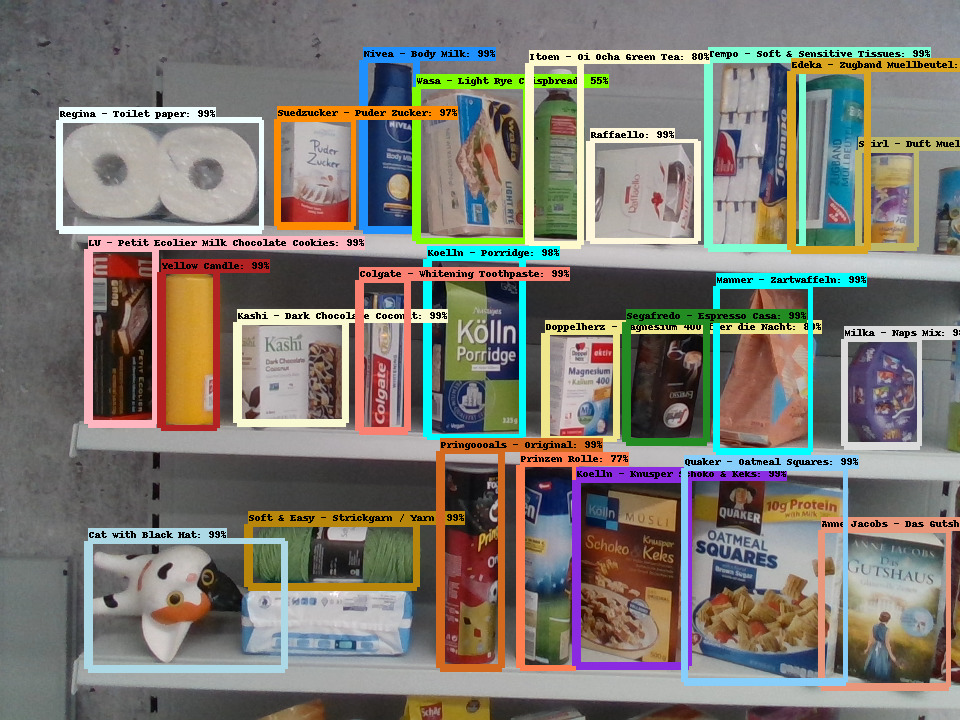} \\
\end{tabular}
\end{center}
\caption{\label{fig:results1} Some results from our real eval dataset: Faster R-CNN trained on our synthetically generated training data robustly detects multiple objects under various poses, heavy background clutter, partial occlusion and illumination changes.}
\end{figure*}

In this section, we report detailed experiments and results underpinning the benefits of our strategy. After describing our experimental setup, we demonstrate that synthetic data generation permits to train state-of-the-art architectures at no cost that outperform models trained on real data. Furthermore, we show through ablation experiments the benefits of curriculum vs random pose generation, the effects of relative scale of background objects with respect to foreground objects, the effects of the amount of foreground objects rendered per image, the benefits of using synthetic background objects, and finally the effects of random colors and blur. 

\subsection{3D models}
In all our experiments, we focus on the detection of 64 different instances of foreground objects showing all very different properties in terms of colors, textures (homogeneous color vs. highly textured), 3D shape and materials (reflective vs. non-reflective). As illustrated by Fig.~\ref{fig:objects}, these objects are mostly classical retail objects that can be found in a supermarket. In addition to these objects of interest, we leverage a large set of approximately 15k objects from different application fields such as industrial objects, household objects or toys that are used for composing the background. For each foreground or background object, we generated a textured 3D model using our in-house 3D scanner.

\subsection{Real Training and Evaluation Data}
\label{Sec:TrainingAndEvalData}
In the present work, we performed all our real data acquisitions using the Intel Realsense D435 camera. While this camera permits to capture RGB and depth images, we focus on RGB only. Using this camera, we built a training and evaluation benchmark of 1158 and 250 real RGB images, respectively, at a resolution of 960x720. Our benchmark training set consists of images picturing random subsets of the objects of interest disposed on cluttered background and in different lighting conditions (natural day/evening light vs. artificial light). The evaluation set consists of images displaying the objects of interest randomly distributed in shelves, boxes or layed out over random clutter. Since it is crucial for reliable object detection, we made sure that in both sets each object is shown in various poses and appears equally (roughly around 120 times for each object in the training set and around 40 times in the evaluation set). All those images were labeled by human annotators and additionally controlled by another observer to ensure highest label quality. This step permitted to correct around 10\% of mislabeled examples which is crucial for fair comparison with synthetic data benefiting from noise-free labels. The amount of time spent for acquiring the real images was around 10 hours and labeling required approximately 185 hours for the training set, with 6 additional hours spent for correction. Note that for real data, acquisition and annotation efforts are always required if new objects are added to the dataset, and images mixing the new objects and the legacy objects need to be generated. In contrast, time spent for scanning the 64 foreground objects was roughly 5 hours, and this is a one time effort: if new objects are added to the dataset, only one scan per additional object is required.

\subsection{Network Architecture}
Modern state-of-the-art object detection models consist of a feature extractor that aims at projecting images from the raw pixel space into a multi-channel feature space and multiple heads that tackle different aspect of the detection problems, such as bounding box regression and classification. In the present work, we use the popular Faster R-CNN ~\cite{faster_rcnn} architecture with an Inception ResNet feature extractor~\cite{inception_resnet}. Weights of the feature extractor have been pre-trained on the ImageNet dataset. Our implementation uses Google's publicly available open source implementation of Faster R-CNN~\cite{Huang17}.

\subsection{Synthetic vs. Real Experiments}
\begin{figure}
  \includegraphics[width=\linewidth]{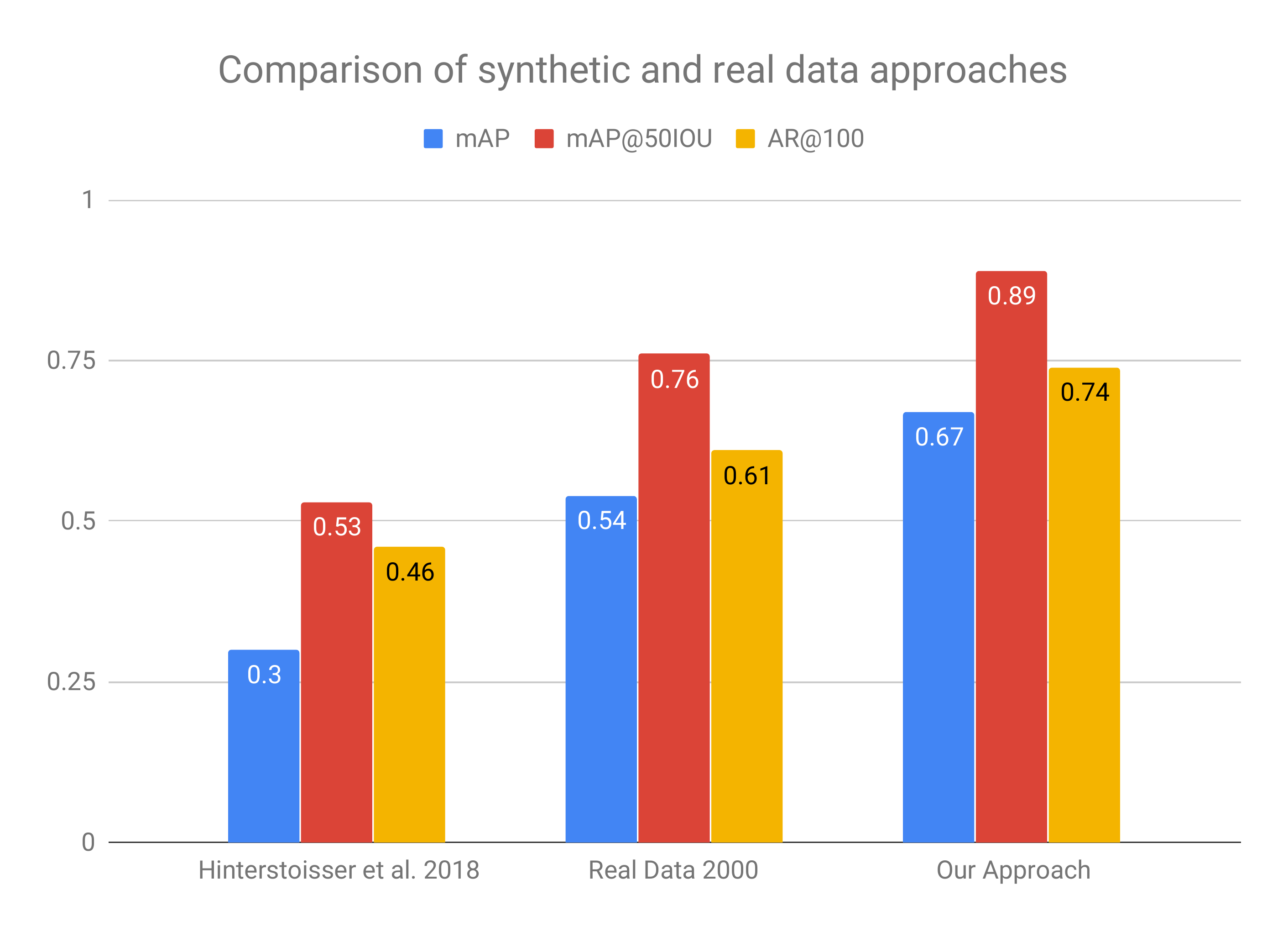}
  \caption{We compare our method with Faster R-CNN trained on the real benchmark training data (see Sec.~\ref{Sec:TrainingAndEvalData}) and with the approach of \cite{Hinterstoisser18}. All models have been trained for the 64 objects of our dataset and tested on the real evaluation dataset (see Sec.~\ref{Sec:TrainingAndEvalData}). Our approach outperforms the other two.}
  \label{fig:realsyntheticresults}
\end{figure}
In this experiment, we aim at demonstrating that our synthetic data generation approach permits to train models that suffer less from the domain gap. To underpin this hypothesis, we compare three Faster R-CNN models initialized using the same weights, the first one being trained according to \cite{Hinterstoisser18}, the second using real data and data augmentation and the third one using our synthetic generation pipeline. All three models have been trained using distributed asynchronous stochastic gradient descent with a learning rate of 0.0001 for 850K iterations. Fig.~\ref{fig:realsyntheticresults} shows the performance of the models in terms of mean average precision (mAP in blue), mean average precision at 50\% intersection over union between ground truth and detected boxes (mAP@50IOU in red) and average recall at 100 detection candidates (AR@100 in yellow). These results clearly demonstrate the benefits of our approach that permits to outperform a model trained on real data in terms of mean average precision as well as average recall. 

\subsection{Ablation Experiments}
In the following experiments, we highlight the benefits of our curriculum learning strategy and investigate the effects of relative scale of background objects with respect to foreground objects, the effects of the amount of foreground objects rendered per image, the influence of the background composition and finally the effects of random colors and blur. As in the previous experiments, models are trained using distributed asynchronous stochastic gradient descent with a learning rate of 0.0001.

\subsubsection{Curriculum vs. Random Training}
\begin{figure}
  \includegraphics[width=\linewidth]{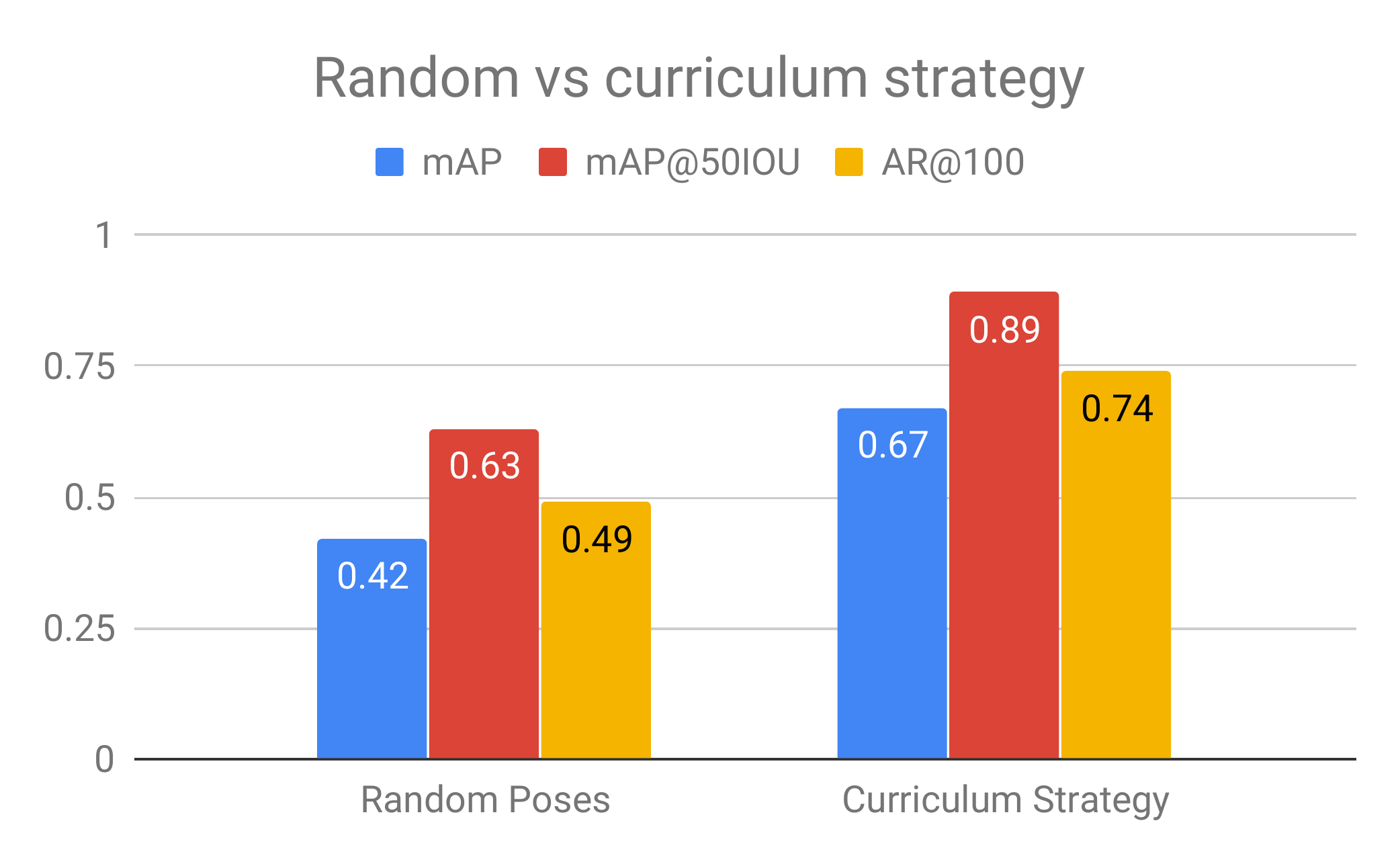}
  \caption{Effect of curriculum strategy vs random poses. Curriculum strategy significantly outperforms random pose generation.}
  \label{fig:curriculum}
\end{figure}
As described in the methods section \ref{sec_fg_layer}, data are generated following a curriculum that ensures that all models are presented to the model equally under pose and conditions with increasing complexity. In this experiment, we compare 2 Faster R-CNN models initialized with the same weights, the first being trained using complete random pose sampling, and the other one following our curriculum strategy. Fig.~\ref{fig:curriculum} clearly shows the benefits of our approach versus naive random sampling strategy.

\subsubsection{Relative Scale of Background Objects}
\begin{figure}
  \includegraphics[width=\linewidth]{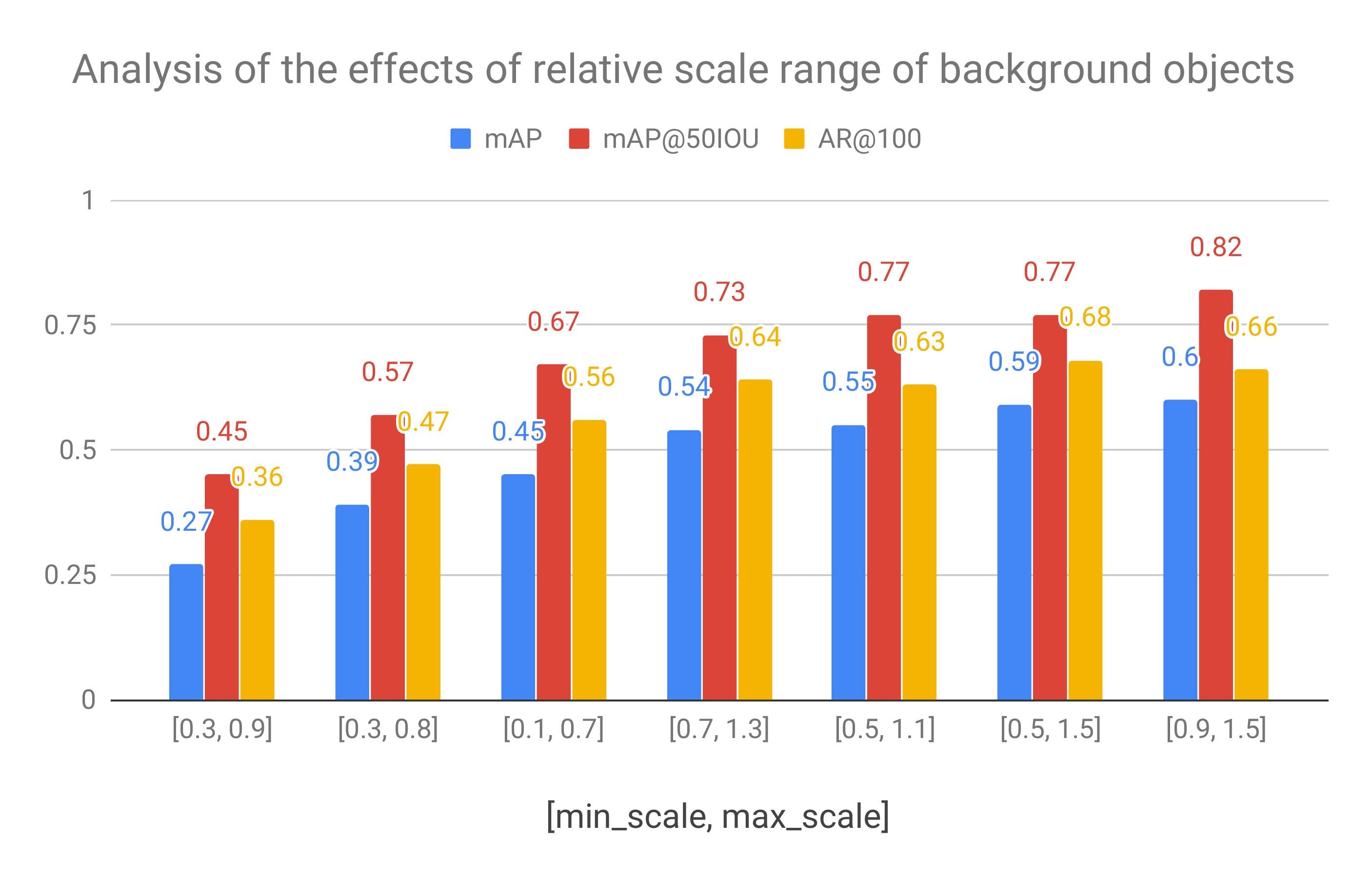}
  \caption{Comparison between models trained using different relative scale ranges for background objects. 
  As we see, properties of the background clutter significantly influences the detection performance.}
  \label{fig:scalerange}
\end{figure}
In the following experiments, we analyze the effects of varying the relative scale range of background objects with respect to foreground objects. Fig.~\ref{fig:scalerange} shows that best results can be obtained for a range that yields background objects of similar or larger size than foreground objects. Using smaller scale ranges yields background images that look more like textures, making it easier for the network to distinguish the foreground objects.

\subsubsection{Amount of Rendered Foreground Objects}
\begin{figure}
  \includegraphics[width=\linewidth]{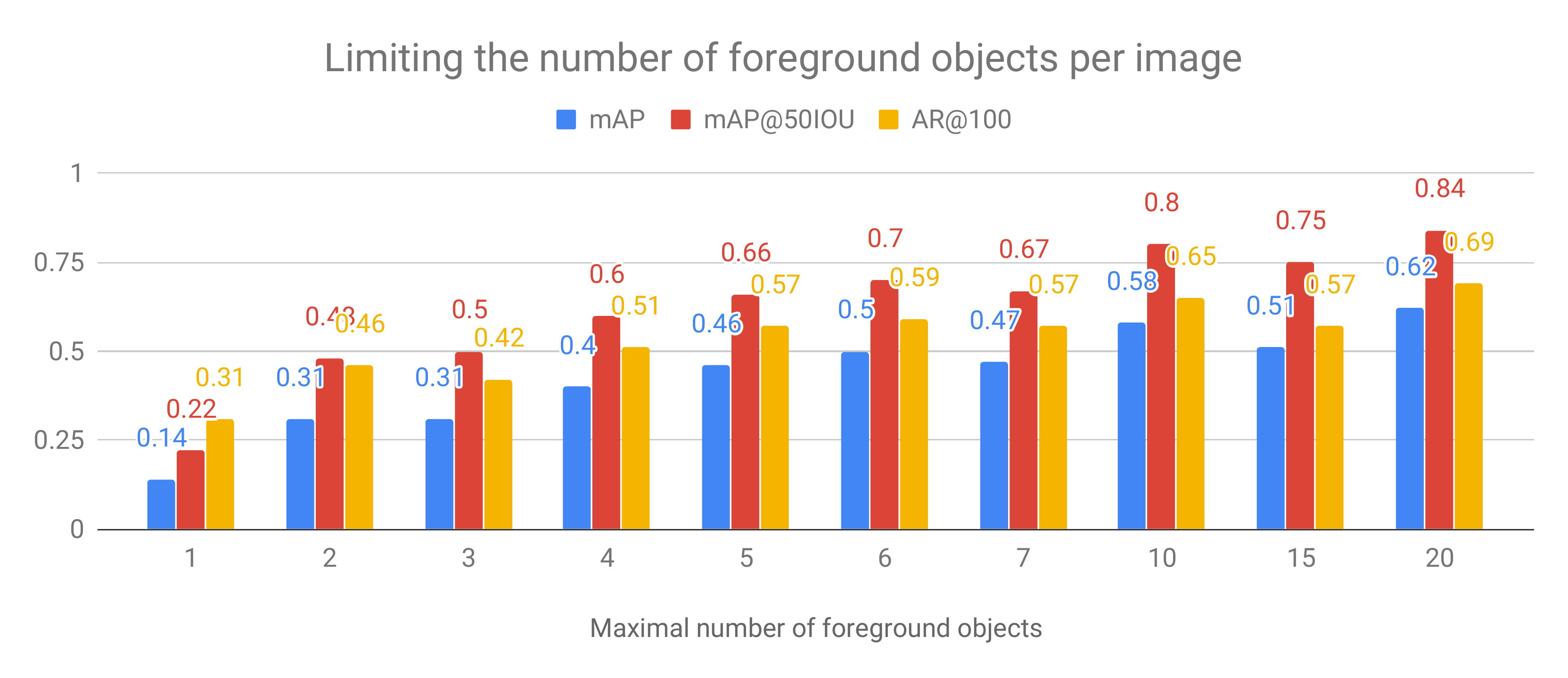}
  \caption{Effect of limiting the number of foreground objects in one image. Detection performance increases with the number of foreground objects rendered in one training image.}
  \label{fig:foregroundresults}
\end{figure}

In this experiment, we study the influence of the amount of foreground objects rendered in the training images. Fig. ~\ref{fig:foregroundresults} clearly shows that a higher number of foreground objects yields better performance. Please note that we only set an upper limit to the number of foreground objects drawn in one image, thus, the average number of objects is typically lower. In particular, in the early stages of curriculum learning we can only fit 8-9 objects in one image on average.

\subsection{Effects of Background Composition}
\begin{figure}
  \includegraphics[width=\linewidth]{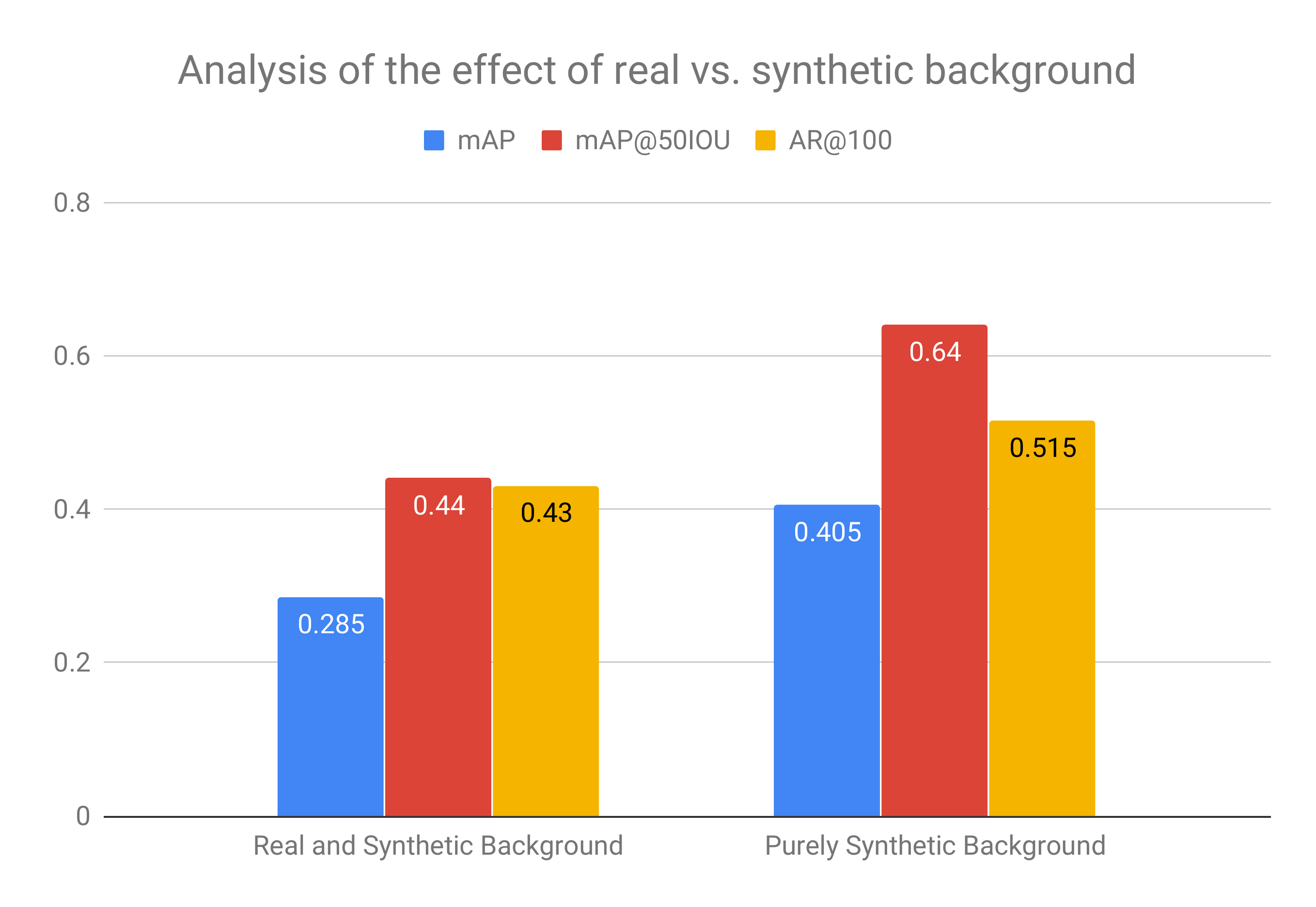}
  \caption{On the left, the model is trained using foreground objects rendered on background images which are partially real and synthetic (as in \cite{Tremblay18,Prakash18}), and on the right, using foreground objects rendered on purely synthesized background images.}
  \label{fig:back_comp}
\end{figure}
In this experiment, we analyze the effect of using purely synthesized background images against real background images which are partially augmented with synthetic objects. To this end, we fix the percentage of the image which is covered by foreground objects ($20\%$ in our case). In the first case, the background is a mixture where $70\%$ of a training sample consists of a real background image and $10\%$ of synthesized background. In the second case, the background consists entirely of synthetically rendered objects. Our results in Fig.~\ref{fig:back_comp} show that the fully synthetic background coverage outperforms images in which only parts of the image are covered by synthetic objects.

\subsubsection{Further Ablation Experiments}
\begin{figure}
  \includegraphics[width=\linewidth]{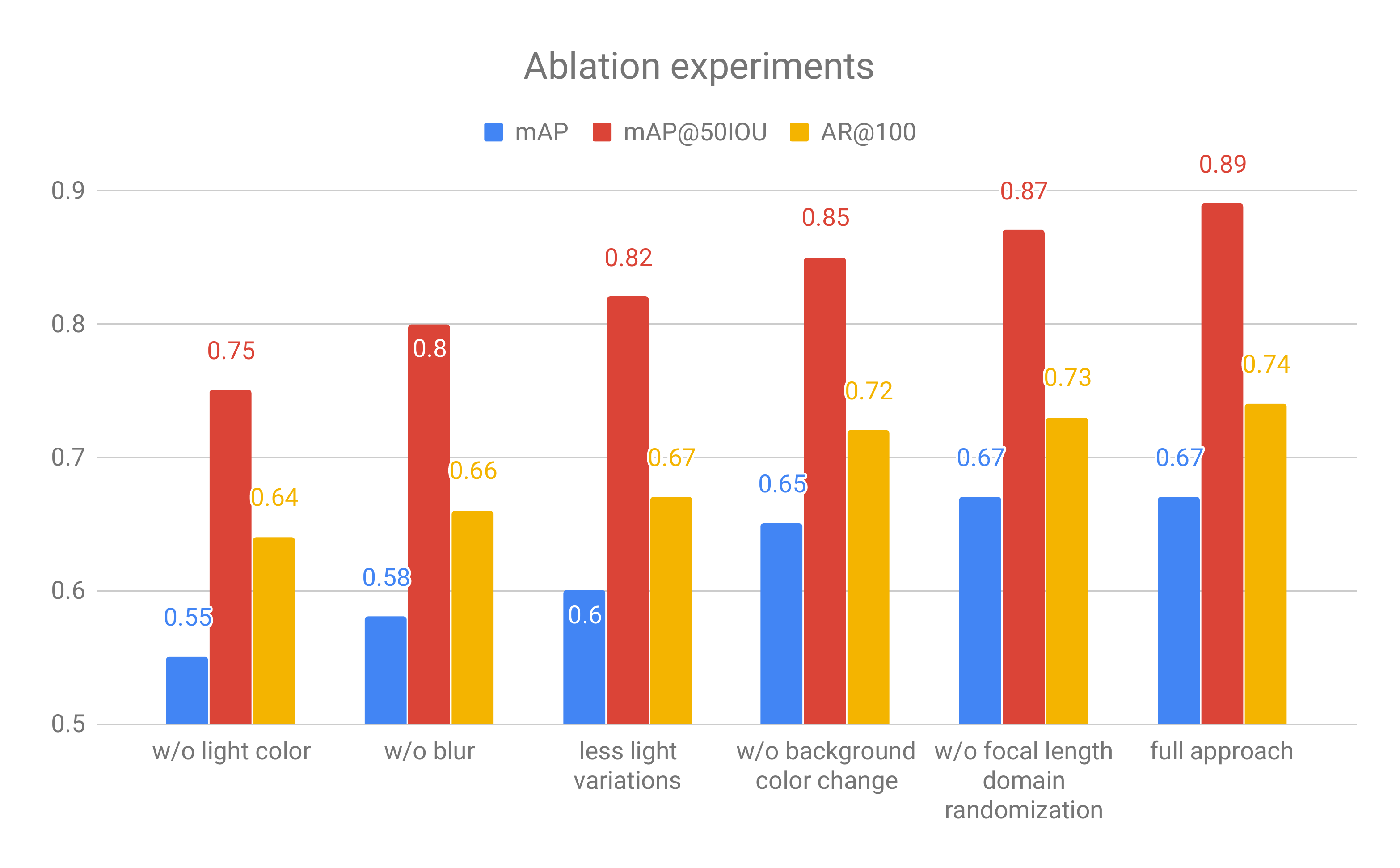}
  \caption{Influences of the different building blocks of our rendering pipeline. Blurring and random light color are important yet simple operations to apply to the synthetic images to improve the results. }
  \label{fig:ablation}
\end{figure}
In the experiments displayed in Fig.~\ref{fig:ablation}, we investigated the influence of the single steps in  the image generation  pipeline. We found that blurring and random light color are most influential, followed by allowing less random light color variations. Randomly varying the focal length of the camera is least important.

\section{Discussion}
We would like to emphasize the main benefits of fully synthetic approaches for object detection.
Consider an object detection system deployed in a warehouse.
They need to maintain a catalogue of thousands of consumer products changing at a high frequency.
While the annotation of large collections of products is itself very costly, the constant updating of this training data,
as a result of changing catalogues, amplifies this issue even more and makes it infeasible to scale.
On the other hand, 3D models often exist during the product design phase or can be easily acquired with off-the-shelf 3D scanners. 
For these reasons, we strongly believe that fully-synthetic data generation approaches are critical for making the deployment and maintenance of large scale object detection pipelines tractable in fast changing real-world environments.
\section{Conclusion}
In this work, we leverage foreground and background 3D models for generating synthetic training data for object detection. 
We introduce a generation and rendering process that follows a curriculum strategy to ensure that all objects of interest are presented to the network equally under all possible poses and conditions with increasing complexity. 
Furthermore, we experimentally demonstrate that models trained in the synthetic domain compare favorably to models trained with synthetic and real data. 
Finally, we show that our approach yields models outperforming object detectors trained purely on real data.

In future work, we will investigate the applicability of our approach for instance segmentation and pose estimation where collecting annotations becomes even more difficult. 

\comment{Furthermore, we will further extend our curriculum strategy process with respect to object-specific geometric and appearance properties, so that we can selectively render more data from objects under specific conditions that are challenging to detect for the network in a given state.}


{\small
\bibliographystyle{ieee}
\bibliography{./string,./cleaned_vision}
}

\end{document}